\documentclass[letterpaper, 10 pt, conference]{ieeeconf}  

\usepackage[varg]{txfonts}
\usepackage{hyperref}
\usepackage{url}

\usepackage{makecell}

\usepackage{subcaption}
\usepackage{graphicx}
\usepackage{lipsum}  
\usepackage{xcolor}
\usepackage{pifont}
\graphicspath{ {images/} }
\usepackage{orcidlink}

\usepackage[backend=biber, style=ieee, doi=true, url=false]{biblatex}
\hypersetup{colorlinks=true,citecolor=blue,urlcolor=blue,linkcolor=blue}

\title{Presenting DiaData for Research on Type 1 Diabetes}
\author{
  Beyza Cinar \orcidlink{0009-0006-3617-0134} \\ 
  Helmut Schmidt University, Germany\\
  \texttt{cinarb@hsu-hh.de} \\ 
   \and
   Maria Maleshkova \orcidlink{0000-0003-3458-4748} \\
   Helmut Schmidt University, Germany\\
   \texttt{maleshkm@hsu-hh.de}
}

\begin{document}

\maketitle

%

\begin{abstract}
Type 1 diabetes (T1D) is an autoimmune disorder that leads to the destruction of insulin-producing cells, resulting in insulin deficiency, as to why the affected individuals depend on external insulin injections. However, insulin can decrease blood glucose levels and can cause hypoglycemia. Hypoglycemia is a severe event of low blood glucose levels ($\le70$ mg/dL) with dangerous side effects of dizziness, coma, or death. Data analysis can significantly enhance diabetes care by identifying personal patterns and trends leading to adverse events. Especially, machine learning (ML) models can predict glucose levels and provide early alarms. However, diabetes and hypoglycemia research is limited by the unavailability of large datasets. Thus, this work systematically integrates 15 datasets to provide a large database of 2510 subjects with glucose measurements recorded every 5 minutes. In total, 149 million measurements are included, of which 4\% represent values in the hypoglycemic range. Moreover, two sub-databases are extracted. Sub-database I includes demographics, and sub-database II includes heart rate data. The integrated dataset provides an equal distribution of sex and different age levels. As a further contribution, data quality is assessed, revealing that data imbalance and missing values present a significant challenge. Moreover, a correlation study on glucose levels and heart rate data is conducted, showing a relation between 15 and 55 minutes before hypoglycemia. 
\end{abstract}
\maketitle
\section{Introduction}
\label{intro}
Type 1 diabetes (T1D) is an autoimmune disorder and one of the most prevalent chronic diseases among children and adolescents, who show worse glycemic control than adults \cite{Laffel2020CITY}. IDF Diabetes Atlas estimates that over 9 million people worldwide were living with T1D in 2024 \cite[p.72]{idf2025}. T1D is associated with insulin deficiency. Therefore, affected patients require lifelong external insulin injections. Patients with T1D usually have irregular and increased glucose levels of more than 180 mg/dL, referred to as hyperglycemia, because glucose cannot be regulated normally. Hyperglycemia can lead to serious complications like diabetic ketoacidosis, and vascular comorbidities \cite{Nilam2011, sergazinov_glucobench_2024}. Diabetes management and therapy includes tailored insulin administration, which, however, has a side effect of decreased glucose levels less than 70 mg/dL, referred to as hypoglycemia \cite{Nilam2011}. Nowadays, biomedical sensors such as continuous glucose measurement (CGM) devices, which measure glucose levels in the interstitial fluid, enhance diabetes therapy and improve glycemic control \cite{sergazinov_glucobench_2024, Vettoretti2020, Laffel2020CITY,aleppo_replace-bg_2017}. In particular, artificial intelligence (AI) methods like machine learning (ML) and deep learning (DL) can be trained on CGM data to predict glucose values or the risk of adverse events (hyperglycemia and hypoglycemia). Notably, neural networks (NNs) report promising performance since they can identify patterns automatically from time series data \cite{Vettoretti2020, Felizardo2021review}. Data analysis can significantly improve tailored diabetes management by discovering individual trends and risks. However, for reliable results, these models need to be trained on large datasets. Usually, models are trained on less than 20 subjects \cite{Felizardo2021review}, which deliver data volumes that are insufficient for most ML approaches but are also under-representative for the patient groups. The limited availability of such datasets remains the leading research gap and hinders the potential of AI-driven approaches.

This study addresses this gap to enhance diabetes care and to enable the training and validation of algorithms that mitigate bias. We present DiaData, a large CGM dataset, by systematically integrating 15 datasets to enable effective training on ML and DL models. In particular, this study makes the following contributions: 1) It provides DiaData, a large dataset with CGM values recorded every 5 minutes (15 datasets). 2) It provides a sub-dataset including demographics of age and sex (10 datasets). 3) It provides a sub-dataset including heart rate data (3 datasets). 4) It investigates the quality of the dataset and explores cohort statistics. 5) Finally, it conducts a correlation study between CGM values in hypoglycemic ranges and heart rate values. 

The remainder of this work is as follows. Section \ref{sec:sota} highlights related work and marks research gaps. Section \ref{sec:methods} summarizes the single datasets used and presents the methods utilized for integrating the dataset. Section \ref{sec:results} assesses data quality and investigates bias in demographics, and conducts a correlation analysis. Findings are discussed in section \ref{sec:discussion}, and lastly, section \ref{sec:conclusion} provides a conclusion.

\section{Related Work}
\label{sec:sota}

In recent years, AI-based approaches have been increasingly explored to enhance diabetes management and prevent adverse events like hypoglycemia. Notably, models for glucose forecasting, insulin optimization, and hypoglycemia prediction have been investigated using various approaches, while neural network-based models are often superior for prediction \cite{Felizardo2021review}. However, a major limitation is the restricted access to public datasets, which hinders the potential of AI models \cite{Felizardo2021review, sergazinov_glucobench_2024}. Without large, representative databases, neither individual nor population-based models can be effectively trained. Moreover, Felizardo et al. state that the majority of studies train on datasets comprising less than 20 subjects, with the OhioT1DM dataset (12 subjects) being the most popular \cite{Felizardo2021review}. Other usually utilized public datasets include the D1NAMO (9 subjects) \cite{dubosson_open_2018}, ShanghaiT1D (25 subjects) \cite{zhao_chinese_2023}, and HUPA-UCM datasets (18 subjects) \cite{hidalgo_hupa-ucm_2024}. Cinar et al. demonstrate that DL models trained on small datasets tend to overfit in hypoglycemia classification. While the OhioT1DM dataset enables short-term prediction, long-term forecasting remains challenging \cite{cinarMaster}. Similarly, smaller datasets like D1NAMO do not support generalized or validated approaches \cite{xie_transfer_2024}. 

In the context of data integration methods, Dong et al. propose an approach focused on big data consisting of three steps. First, schema mapping refers to identifying the mappings between the global and local schemas to define which set of attributes contains the same information. Second, record linkage focuses on linking a static set of structured records that have the same schema. Third, data fusion integrates all data \cite{dong_big_2013, eppig_mouse_2015}. Sourlos et al., focusing on health data, propose to first define the specific use case and clinical target, so that the most suitable features can be selected. Furthermore, the integrated dataset should contain ground truth and labels. Also, fairness and the risk of bias should be reported by providing an initial data analysis on the population and its subgroups. Finally, a benchmark model performance can be provided \cite{sourlos_recommendations_2024}.  

Focusing on diabetes research, two fused datasets collecting glucose were presented. 
First, Arévalo et al. curated a dataset from the MIMIC III database, but did not use CGM device data. They matched glucose and insulin values collected in the intensive care unit from electronic health records \cite{robles_arevalo_data-driven_2021}. 
Second is GlucoBench, which integrates datasets including CGM data. Sergazinov et al. fused five different databases comprising different diabetes types and no diabetes subjects. Datasets were included if they contained at least five subjects, each with not more than 16 hours of consecutive missing CGM data. Also, subjects with poor data quality were removed, and the data were clipped to only relevant ranges between 20 mg/dL to 400 mg/dL. One disadvantage is the imbalance in the data volume, diabetes types, and demographics of the single datasets, which can lead to bias in the integrated dataset. Finally, data quality is improved \cite{sergazinov_glucobench_2024}. While CGM data is included, GlucoBench may not be effective for hypoglycemia prediction due to the limited scale of its T1D subgroup. 

\section{Methodology}
\label{sec:methods}
The absence of large-scale CGM datasets from individuals with T1D in free-living settings represents a key research gap, especially in the context of hypoglycemia prediction.
Therefore, this study systematically integrates 15 different datasets collecting CGM data of patients with T1D. The objective is to improve glucose management and the performance of prediction models. An integrated dataset can be used to train generalized models and to validate models for glucose forecasting. To enable generalization, we aimed to include at least 1000 different subjects and represent different age levels.

Initially, datasets were collected and individually sampled at a uniform frequency of 5 minutes. Then, the datasets were integrated into one main dataset comprising all subjects' IDs, timestamps, and the corresponding glucose measurements. Moreover, two sub-databases were extracted from the main database: 1) sub-database I included demographics of age and sex, and 2) sub-database II included heart rate values. Consequently, three complete datasets are provided, rather than relying on a single dataset including all features but suffering from missing values.

\vspace{1em}
\subsubsection{Dataset Collection}

\label{sec:DataCollection}
The data integration framework of Sourlos et al. and Dong et al. was adapted in this work \cite{sourlos_recommendations_2024, dong_big_2013}. First, the use cases for the database were identified, which are hypoglycemia classification and glucose forecasting. Second, datasets were collected based on the following conditions: 1) They should include CGM data. 2) They should include only patients with T1D. 3) They should be mainly collected under free-living conditions. The final cohort consisted of 15 datasets and 2510 patients, yielding over 800,000 days of data in total. Statistics of the single datasets are presented in Table \ref{tab:datasetoverview}. 
\begin{table*}[ht]
\centering
\caption{Overview of the Single Datasets}
\label{tab:datasetoverview}
\begin{tabular}{|p{2.7cm}|p{1.2cm}|p{1cm}|p{0.9cm}|p{0.9cm}|p{0.9cm}|p{0.9cm}|p{0.9cm}|p{0.9cm}|p{0.8cm}|p{0.6cm}|}
\hline
\textbf{Data} & \textbf{Subjects} & 
\textbf{Min Values} & \textbf{Max Values} &\textbf{Mean Values} & \textbf{Min Days} & \textbf{Max Days} & \textbf{Mean Days} & \textbf{Missing Values} & \textbf{Sex \& Age} & \textbf{HR}  \\
\hline 
T1DGranada \cite{RodriguezLeon2023}    & 736  &1167 & 389515& 87419  & 6  & 1531 & 443 &29682412& \ding{52} & \ding{55} \\
DiaTrend \cite{DiatrendPaper} & 54 &8075&448006 & 134833 & 30  & 2179 & 564 &1500336 & \ding{52} & \ding{55} \\
CITY \cite{jab_center_diabetes}  & 153    &2189& 105768& 54004  & 10  & 709  & 348  & 7134271 & \ding{55} & \ding{55} \\
DLCP3 \cite{jab_center_diabetes} & 169   &3253&66589 & 44701  & 11  & 867  & 179 & 1213160 &\ding{52} & \ding{55}\\
HUPA-UCM \cite{HupaMendeley} & 25     &159& 227597& 21265  & 0   & 1152 & 174 &  700448 & \ding{55}&\ding{52}\\
PEDAP \cite{jab_center_diabetes}  & 103   &3687& 118667& 70389  & 13  & 480  & 286 &1253565 &\ding{52} &\ding{55} \\
RBG \cite{jab_center_diabetes}  & 226  &16188&93355 & 65047  & 72  & 820  & 277 &3425489 & \ding{55}  &\ding{55}  \\
RT-CGM \cite{jab_center_diabetes} & 451  &932&95230 & 42809  & 6   & 451  & 355 &26956667&\ding{52} &  \ding{55} \\
SENCE  \cite{jab_center_diabetes} & 144  &2260&108294 & 57687 & 11  & 419  & 342 & 5928255 &\ding{52} & \ding{55} \\
SHD \cite{jab_center_diabetes} & 200   &431&5118& 3220   & 5   & 58   & 14 &253575& \ding{52} &  \ding{55} \\
WISDM \cite{jab_center_diabetes}  & 206  &3604& 109474& 51307  & 12  & 496  & 356 &10644507 &\ding{52} & \ding{55} \\
ShanghaiT1D \cite{ShanghaiFigshare} & 16   &1069&4015 & 2940   & 3   & 13   & 9 &  5&\ding{52} & \ding{55} \\
D1NAMO \cite{D1NAMOZenodo} & 9      &115&1407 & 884    & 0   & 4    & 3  & 1338 & \ding{55}&\ding{52} \\
DDATSHR \cite{ictinnovaties_diabetes_heart_rate_2025} & 18    &79&24349 & 16011  & 15  & 87   & 72 &  89820 &\ding{52}& \ding{52}  \\
T1GDUJA \cite{T1GDUJAZenodo} & 1      &41292&41292 & 41292  & 246 & 246  & 246 &  29717& \ding{55} & \ding{55} \\
\hline
\end{tabular}
\end{table*}
Differences in data availability and measured glucose values can be observed, with some subjects having limited records. This imbalance can introduce a risk of bias. The number of patients per dataset ranges from a minimum of 1 to a maximum of 736. Notably, T1DiabetesGranada (T1DGranada) and RT-CGM datasets include more subjects than the average, whereas T1GDUJA, DDATSHR (dataset diabetes adolescents time-series with heart rate), D1NAMO, and ShanghaiT1D contain fewer. 

\textbf{Description of Datasets:} In the following, the single datasets will be summarized. In most of the datasets, sex was equally distributed. 

The T1DGranada dataset comprises CGM data, demographic variables (sex and birth year), biochemical measurements, and clinical parameters. Participants' ages ranged from 12 to 81 years, with a mean of 40 years. In total, over 257,000 days of CGM data were collected, with some individuals recording data over 4 years. CGM values were measured with the FreeStyle Libre 1 and 2 devices from Abbott, which have a frequency of 15 minutes. Biochemical parameters were collected at follow-up visits conducted every 3 or 6 months~\cite{RodriguezLeon2023}. 

The DiaTrend dataset provides over 27,000 days of CGM data and more than 8,000 days of insulin pump data collected from two independent cohorts. The first cohort included 17 patients aged 25 to 74 years, who all used CGM devices and insulin pumps. The second cohort consisted of 37 participants from the SweetGoals study, aged 19 to 25 years, each with over 18 months of CGM data. Various CGM systems were employed across the study, including devices from Dexcom, Abbott, and Medtronic. In total, the dataset includes 17 male and 37 female participants. Additionally, demographic and clinical information such as sex, race, and HbA1c values was collected for each subject~\cite{DiatrendPaper}. 

The HUPA-UCM dataset includes data on CGM, insulin, physical activity, heart rate, and sleep quality and quantity. CGM data were collected using the FreeStyle Libre 2 system, while wearable data were obtained via Fitbit Ionic smartwatches. Participants on multiple daily insulin (MDI) therapy recorded basal and bolus insulin doses, as well as carbohydrate intake. Physiological data like heart rate, energy expenditure, steps, and sleep were collected by the Fitbit device under free-living conditions. The cohort had a mean age of 39 years, a mean HbA1c value of 7.37\%, and a mean diabetes duration of 17 years~\cite{hidalgo_hupa-ucm_2024}.

The ShanghaiT1D dataset comprises multimodal data collected under free-living conditions from adults for 3 to 14 days. Data includes CGM measurements, clinical characteristics, laboratory values, medication usage, and dietary records. Participants wore the FreeStyle Libre from Abbott, which recorded glucose levels at 15-minute intervals. A total of approximately 15,000 CGM data points were collected. Participants had a mean age of 58 years. The duration of diabetes ranged from 2 to 16 years. From medical records, information on diabetes diagnosis and treatment, comorbidities, medications, and laboratory results from the last 6 months was extracted. Laboratory results include markers of the glucose metabolism, lipid profiles, and renal function. Moreover, a physical examination was conducted to measure height and weight. Daily dietary intake and meal times were self-reported. Recorded medications include hypoglycemic medications like oral agents and insulin~\cite{zhao_chinese_2023}.

The D1NAMO dataset comprises 3 to 4 days of CGM data collected from adults under real-life conditions. Also, physiological and behavioral data were recorded using wearable devices. These include accelerometer data for physical activity, ECG for cardiac activity, respiratory measurements, and annotated food images. CGM measurements were obtained using the iPro2 Professional CGM sensor, recording glucose values every 5 minutes. Physiological signals were collected with the Zephyr BioHarness 3. Participants were required to be at least 18 years old, with ages ranging from 20 to 79 years. On average, approximately 50 hours of signal data were recorded per subject. In total, 8,000 CGM measurements across the cohort are included. 6 subjects were male and 3 were female~\cite{dubosson_open_2018}.

The T1GDUJA dataset involves data from only one subject, who had data collected for 226 days~\cite{T1GDUJAZenodo}. 

The CITY, DLCP3, PEDAP, ReplaceBG (RBG), RT-CGM, and WISDM datasets originate from independent multicenter randomized clinical trials. 

The CITY dataset involves patients aged 14–24 years and compares treatment with CGM with conventional blood glucose monitoring (BGM). The CGM group included 74 participants, and the BGM group 79. The overall mean age is 17 years. The mean diabetes duration was 9-10 years, considering both groups. Insulin pump use was reported in 49\% and 47\% of CGM and BGM participants, respectively. HbA1c levels in both groups were measured at baseline, 13, and 26 weeks, reporting a value of 8.9\% at baseline in both groups, which declined to 8.5\% in the CGM group at 26 weeks. The Dexcom G5 CGM, sampling glucose every 5 minutes, was used~\cite{Laffel2020CITY}. 


The PEDAP dataset originates from a 13-week clinical trial involving children aged 2 to 6 years. The trial investigated the efficiency of closed-loop systems. Therefore, children were randomly assigned to a closed-loop insulin delivery system (68 subjects) or standard care (34 subjects), which included treatment via an insulin pump or MDI combined with CGM. Participants had baseline HbA1c levels ranging from 5.2\% to 11.5\%, and a mean age of 4 years. The case group had a mean HbA1c score of 7.5\%, and the control group of 7.7\%. Sex distribution was balanced in the case group, while 56\% of the control group participants were female. The duration of diabetes among participants ranged from 6 months to 5 years. Glucose values were continuously monitored using the Dexcom G6 system. The HbA1c levels were collected at baseline and week 13~\cite{Wadwa2023}. 

This RGB dataset was collected during a 6-month clinical trial assessing the safety and efficacy of using CGM without the addition of BGM compared to combining CGM and BGM. 149 subjects were assigned to the case group and 77 to the control group. Patients were at least 18 years old. The CGM device used was the Dexcom G4 Platinum with the enhanced 505 algorithm, recording glucose values at 5-minute intervals. HbA1c values were collected at baseline, week 13, and week 26. In the case group, participant's age ranged from 19 to 78 years, with a mean of 44 years, and a diabetes duration ranging from 2 to 64 years, with a mean of 23 years. In the control group, age ranged from 25 to 69 years, with a mean of 45 years, and diabetes duration from 4 to 58 years, with a mean of 25 years~\cite{aleppo_replace-bg_2017}.

In the RT-CGM dataset, a clinical trial was conducted to investigate the efficiency of CGM devices. Participants were assigned to either a CGM intervention group (232 subjects) or a control group (219 subjects). In the CGM group, 122 subjects were at least 18 years old, and 110 were less than 18, while in the control group, 106 were at least 18, and 113 were at least 18. Both groups completed standardized questionnaires at baseline and at 26 weeks~\cite{RTC2010}. 

The SENCE dataset contains CGM data from children aged 2 to under 8 years. The trial evaluated the efficacy and safety of CGM use in young children using the Dexcom G4 Platinum CGM System with the enhanced 505 software algorithm. Glucose levels were recorded every 5 minutes over 14–21 days. Eligibility criteria included a minimum diabetes duration of 3 months, HbA1c levels between 7.0\% and 9.0\% within 30 days before screening, and treatment with an insulin pump or MDI. HbA1c values were assessed following an initial masked CGM period of 14 days. In addition to glycemic data, demographic, clinical, socioeconomic, and educational characteristics, as well as insulin regimen details, were collected~\cite{DiMeglio2020}. 

The WISDM data were collected during a 6-month controlled trial involving adults aged at least 60 years. The primary aim of the trial was to evaluate whether real-time CGM could reduce hypoglycemia and improve quality of life in older adults. Thus, patients were assigned to an intervention group using real-time CGM or a control group receiving standard BGM therapy. Participants were required to have an HbA1c level below 10\% at screening, to use either an insulin pump or MDI injections, and to have not used real-time CGM in the last 3 months before enrollment. CGM data were collected via the Dexcom G4 Platinum CGM system with the enhanced 505 software algorithm. The median age of the cohort was 68 years, with a median diabetes duration of 36 years. The mean HbA1c was 7.5\%. Of the participants, 53\% used insulin pumps. A median of 326 hours of CGM data per participant were collected~\cite{Carlson2019}.

The Severe Hypoglycemia Dataset (SHD) originates from a multicenter case-control study that investigated potential factors contributing to the occurrence of severe hypoglycemia in older adults. Included subjects were aged 60 years or older and had a diabetes duration of at least 20 years. All subjects were required to be on insulin therapy, while exclusion criteria included current use of CGM devices and the presence of significant comorbidities. The study included 101 case subjects who had experienced at least one severe hypoglycemia requiring the assistance of another person within the last year, and 100 control subjects who had not experienced severe hypoglycemia in the past 3 years. The mean age was 68 years in both groups. In the case group, sex was equally distributed, insulin pumps were used by 58\%, and the average HbA1c level was 7.8\%. In the control group, females comprised 44\%, insulin pumps were used by 59\%, and the average HbA1c level was 7.7\%. Each participant underwent two study visits during which physical examinations and a questionnaire were conducted~\cite{Weinstock2015}. 

Finally, the DDATSHR dataset does not specify data characteristics.

\vspace{1em}
\subsubsection{Dataset Sampling} 
\begin{figure*}[h]
\centering
\includegraphics[width=10cm,clip]{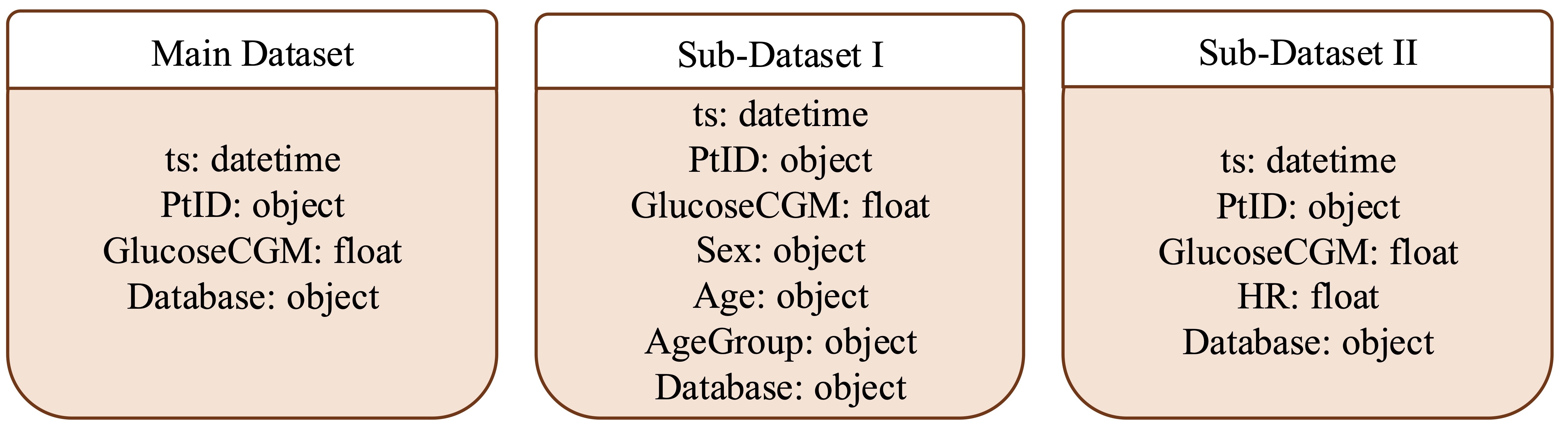}
\caption{Database Schema}
\label{fig:schema}       
\end{figure*}

After the corpus generation, global and local schemas were defined, which can be seen in Figure \ref{fig:schema}. 
To enable seamless integration, local column names were renamed. For each relation $X \in \{X_1, X_2, \dots, X_n\}, \quad \rho_{X(a', b', c', d', e')} \left(\pi_{a, b, c, d, e}(X) \right)$ was applied in which $X$ is the database, (a', b', c', d', e') are (GlucoseCGM, PtID, Sex, Age, HR), (a, b, c, d, e) are the individual old values. 
The single databases were processed separately. 

First, one subject from the HUPA-UCM database had misaligned rows, which were manually corrected. To the first 3 lines in the file, 4 extra semicolons were added at the end of the last column. Second, all datasets were resampled to the same frequency of 5 minutes. To enable resampling, a timestamp column named "ts" was created for each database. For datasets reporting the date and the time as separate columns, the attributes were concatenated and converted to a datetime object. The RBG and SHD datasets only reported the number of days passed since admission and the measured time as separate columns. The reported days were added to a randomly generated date (01.01.2024) to create a valid datetime column. 

The timestamps were rounded to 5 minutes to have uniform timestamps across each subject and database, and resulting duplicates were removed. Then, the timestamps were resampled to 5-minute intervals. Some datasets (DDATSHR and RT-CGM) used multiple sensors. Thus, the measurement frequency was identified before resampling. 
Dataframes including heart rate were only rounded to 1-minute intervals, and duplicates were removed. Thereafter, they were aligned to the glucose measurements in 5-minute intervals with an outer join. In total, 5 datasets had to be undersampled since those were collected with frequencies of 10 or 15 minutes. Undersampling is warranted in this context, as 15-minute gaps can be effectively addressed through imputation \cite{SuwenLin}. Produced missing values between collected parameters were imputed with feedforward filling. Standardizing all data to 15-minute intervals would reduce the data volume. Moreover, the differentiation between 5, 10, and 15 minutes in short-term horizons is essential for glucose forecasting and hypoglycemia classification.
\vspace{1em}
\subsubsection{Dataset Integration} 

Two datasets reported glucose in mmol/L (D1NAMO and DDATSHR). Thus, the glucose values were converted to mg/dL by multiplying by $18.02$. The glucose column of the CITY and D1NAMO datasets included CGM values and manually estimated glucose values. These were split into two distinct columns. The DDATSHR and HUPA-UCM datasets also store historic and scan glucose values. For the integration into a single column before resampling, the historical values were substituted with their corresponding, time-aligned scan glucose readings. 

Then, demographics (age and sex) were extracted from the corresponding dataframes. For the T1DGranada dataset, the age was computed from the reported birthyear of the patient and the timestamps. The SHD dataset mentioned that patients at least aged 60 were chosen for the study. Thus, the age was stored as a string of "60-100", assuming 100 to be the maximum age. All extracted values were aligned to the timestamps and patient IDs. For all databases, sex was stored as "F" for female and "M" for male. Finally, each database was added a "Database" column.

After having a semantic equality among the datasets, they were matched adhering to the defined global schema, and the data were fused. While integrating the datasets, a column for "AgeGroups" was created since some ages were reported as strings in ranges. Based on the saved age ranges, bins of (0$-$2, 3$-$6, 7$-$10, 11$-$13, 14$-$17, 18$-$25, 26$-$35, 36$-$55, 56$+$) were defined. Finally, patients having only null values for the "GlucoseCGM" column were removed from the integrated dataset.

\section{Quality Assessment}
\label{sec:results}

This section presents the quality assessment of the integration framework. Therefore, the glucose characteristics of the databases and bias are analyzed. From Table~\ref{tab:datasetoverview}, it was already observed that some subjects have collected a larger data volume than average, which could introduce a bias risk. Databases T1DGranada and RT-CGM include more subjects than the average, while T1GDUJA, DDATSHR, D1NAMO, and ShanghaiT1D have significantly fewer.   

The main database integrates all 15 datasets, while sub-dataframe I includes 10 (RT-CGM, T1DGranada, DiaTrend, DLCP3, PEDAP, SENCE, SHD, WISDM, ShanghaiT1D, DDATSHR) and sub-dataframe II includes 3 (HUPA-UCM, DDATSHR, D1NAMO) datasets. This distribution underlines the need for more multivariate datasets, including vital parameters. Table~\ref{tab:datasetfusedsummary} summarizes the fused datasets. Combining all datasets, each person has on overage one year of CGM data collected and 149 million usable CGM points without missing values. Sub-dataset I has similar characteristics. Sub-dataset II represents only 0.5\% of the total data in the main dataset. Therefore, even if training models with multiple wearable data increases performance~\cite{Felizardo2021review}, sub-dataset II may perform worse. 
\begin{table*}[ht]
\centering
\caption{Summary of Integrated Datasets}
\label{tab:datasetfusedsummary}
\begin{tabular}{|l|r|r|r|r|r|r|r|r|r|}
\hline
\textbf{Dataset} & \textbf{Subjects} & \textbf{Total Values} & \textbf{Values $\le54$} & \textbf{Values $\le70$}  & \textbf{Values $\ge180$} & \textbf{Values $70-180$} & \textbf{Mean days}  & \textbf{Mean values}\\
\hline
Main Dataset & 2510 & 149112702 & 1588956& 4606683 & 55934561 & 86982502&328 &  59407   \\
Sub-Dataset I  & 2096 &125585996 &1354676  & 3972567 &46249692 & 74009061&335 &  59916  \\
Sub-Dataset II & 51   & 806547 & 10517 & 30767 &194886 & 570377 & 108 & 15814 \\
\hline
\end{tabular}
\end{table*}

\subsection{Glucose Characteristics}

\begin{figure}[ht]
\centering
\begin{subfigure}{0.4\textwidth}
    \includegraphics[width=\textwidth]{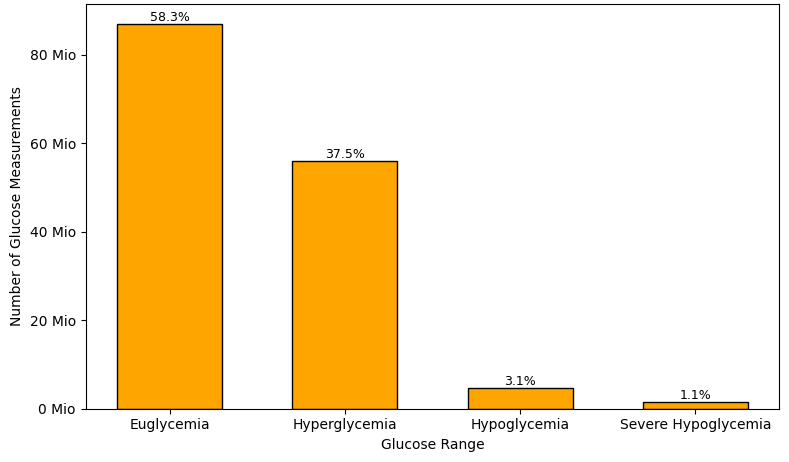}
    \caption{Distribution of Glucose Values}
    \label{fig:glucdistr}
\end{subfigure}
\begin{subfigure}{0.4\textwidth}
    \includegraphics[width=\textwidth]{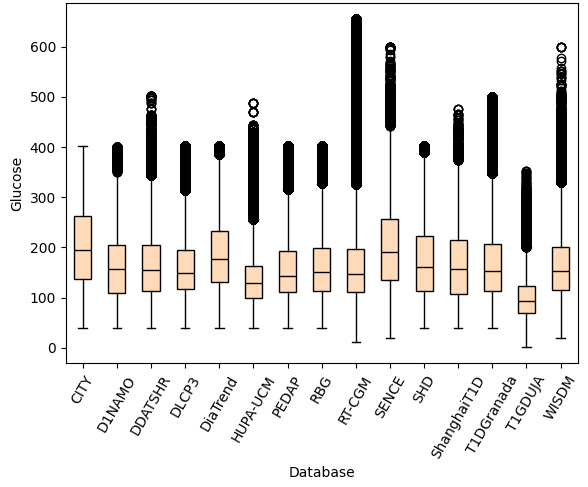}
    \caption{Variance of Glucose Levels}
    \label{fig:singleglucbox}
\end{subfigure}
\caption{Analysis of Glucose Levels in the Main Dataset}
\label{fig:glucchar}
\end{figure}

The main dataset comprises 6.2 million hypoglycemic, 55.9 million hyperglycemic, and 86.9 million target range (euglycemic) data points. This introduces a significant imbalance, particularly for hypoglycemia, which makes up only 4\% of all collected glucose values. Moreover, the distribution between hypoglycemia and severe hypoglycemia is imbalanced, as visualized in Figure \ref{fig:glucdistr}. Thus, the prediction of hypoglycemia could be more challenging using all data. In the context of hypoglycemia classification restricted to hypoglycemic data points, a total of 21,000 days are available from 2501 subjects, excluding missing values. In contrast, hyperglycemia is well represented with 37\%. This reveals the challenge of maintaining euglycemia. However, prolonged and frequent hyperglycemia is associated with vascular comorbidities and should be minimized~\cite{Nilam2011}.

Figure \ref{fig:singleglucbox} illustrates the boxplots of glucose ranges across the single datasets. In particular, a lot of outliers are detected, indicating artifacts in CGM device measurements between 400-700 mg/dL. Therefore, the confidence rate of the CGM devices needs to be investigated to filter outliers. Similarly, physiologically implausible levels like very low glucose values of zeroes should be filtered. Furthermore, significantly elevated glucose levels are noted in the RT-CGM, SENCE, and WISDM datasets, and the least amount of outliers can be seen in the CITY, SHD, and DiaTrend datasets. In most datasets, the third quartile predominantly comprises hyperglycemic data points, and the whiskers to the maximum values are longer than to the minimum, underscoring the prior findings. It can be seen that the median is not always centered and that the data is skewed toward hyperglycemia. If not centered, the median is closer to the lower values.
In addition, across the datasets, a difference in the length of boxes indicates variation between the single datasets, which could depend on age or the study settings. In particular, the T1GDUJA and HUPA-UCM datasets show less variance compared to the remaining datasets, demonstrating that the subjects experience less glucose variation. While values in the HUPA-UCM dataset are typically in the euglycemic range, they behave in the lower values in the T1GDUJA dataset. In contrast, the CITY, SENCE, and SHD datasets present more variance, indicating larger glucose ranges, of which most behave in hyperglycemic ranges. Still, the majority of datasets are mainly in the euglycemic range. These observations conclude that DiaData encompasses a diverse range of subject characteristics, which could improve its generalization performance. However, proper preprocessing is essential before using the data for training.

\subsection{Missing Values}

\begin{figure}[ht]
\centering
    \includegraphics[scale=0.35]{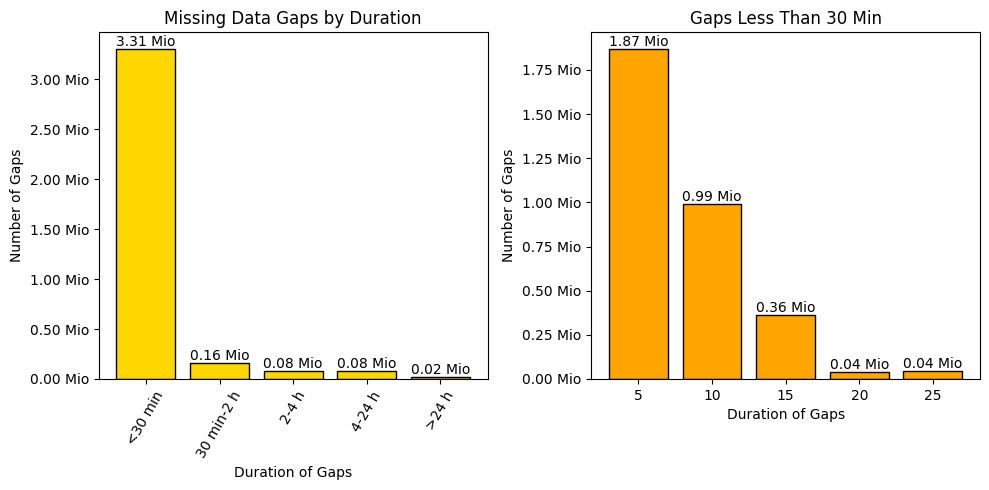}
    \caption{Missing Value Count of the Main Dataset \label{fig:missingmain}}
\end{figure}

Missing values can occur due to sensor errors, behavioral factors, and not wearing the wearable device continuously. Examining the quality of the dataset from Figure \ref{fig:missingmain}, it can be seen that missing values are a significant problem with the database. Most studies probably have not measured data continuously during the whole study duration across all subjects. In total, up to 88.8 million glucose values are missing. As summarized in Table \ref{tab:datasetoverview}, the datasets holding most of the subjects have the majority of missing values (T1DGranada and RT-CGM). Examining the missingness of values in more detail, Figure \ref{fig:missingmain} demonstrates that the majority of gaps occur between 10 and 30 minutes, indicating sensor or environmental errors. Most of these are 5-minute gaps which can be imputed with adequate methods. Thus, even if lots of missing values exist, proper preprocessing can improve data quality, for instance, with linear imputation, which performs well for short-term gaps \cite{SuwenLin}. In contrast, larger gaps of 4 to more than 24 h should probably be removed. 

\subsection{Demographics}

\begin{figure*}
\centering
\begin{subfigure}{0.35\textwidth}
    \includegraphics[width=\textwidth]{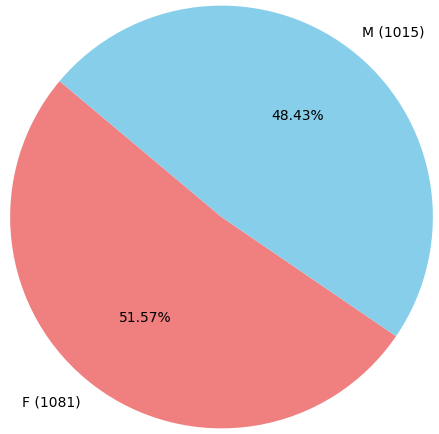}
    \caption{Sex Distribution}
    \label{fig:glucagesex}
\end{subfigure} 
\begin{subfigure}{0.35\textwidth}
    \includegraphics[width=\textwidth]{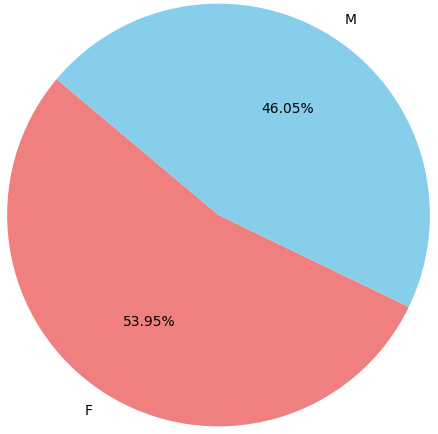}
    \caption{Measurement Frequency by Sex}
    \label{fig:glucagesexfreq}
\end{subfigure} 
\begin{subfigure}{0.35\textwidth}
    \includegraphics[width=\textwidth]{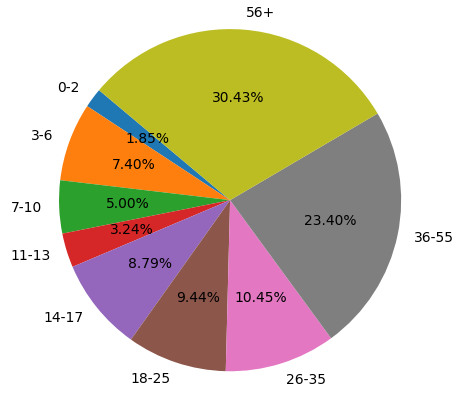}
    \caption{Female Ratio by Age Group}
    \label{fig:sexagef}
\end{subfigure}
\begin{subfigure}{0.35\textwidth}
    \includegraphics[width=\textwidth]{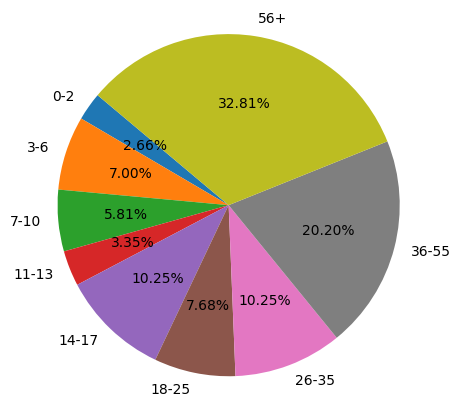}
    \caption{Male Ratio by Age Group}
    \label{fig:sexagem}
\end{subfigure}
\begin{subfigure}{0.35\textwidth}
    \includegraphics[width=\textwidth]{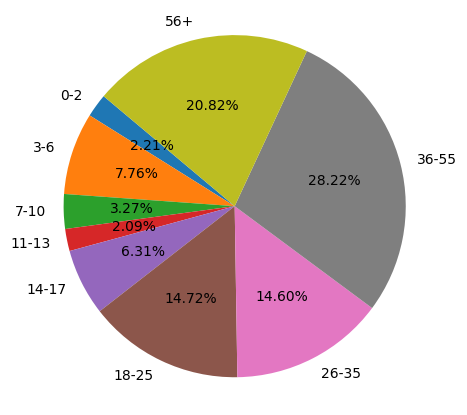}
    \caption{Measurement Frequency of Females by Age Group}
    \label{fig:glucagesexf}
\end{subfigure} 
\begin{subfigure}{0.35\textwidth}
    \includegraphics[width=\textwidth]{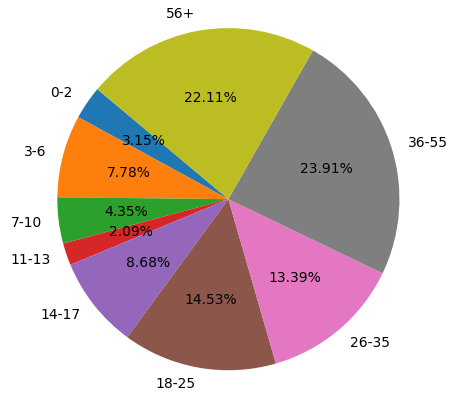}
    \caption{Measurement Frequency of Males by Age Group}
    \label{fig:glucagesexm}
\end{subfigure} 
\caption{Distribution of Demography}
\label{fig:agedist}
\end{figure*}

\begin{figure}
\centering
\begin{subfigure}{0.4\textwidth}
    \includegraphics[width=\textwidth]{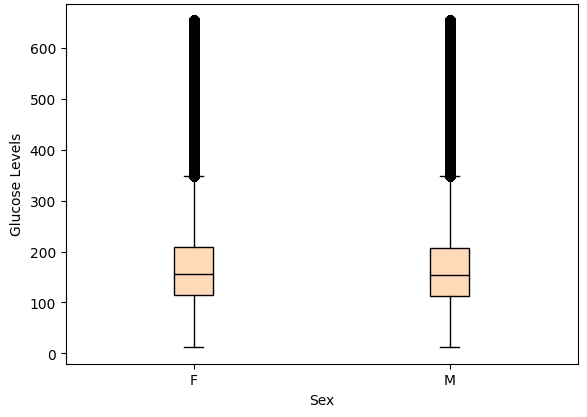}
    \caption{Variance of Glucose Across Sex}
    \label{fig:sexbox}
\end{subfigure}
\begin{subfigure}{0.4\textwidth}
    \includegraphics[width=\textwidth]{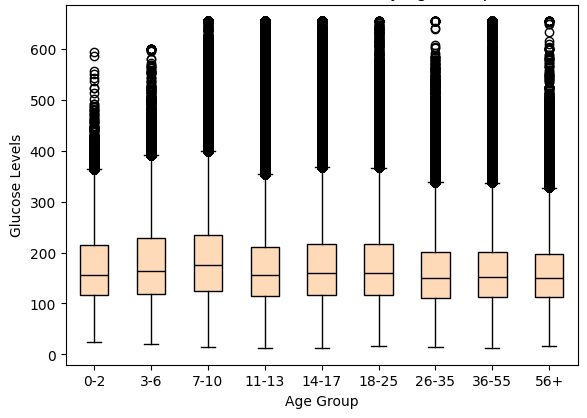}
    \caption{Variance of Glucose Across Age Groups}
    \label{fig:agebox}
\end{subfigure}
\caption{Boxplot Representation of Glucose Levels Across Demographic Groups}
\end{figure}

Turning now to the analysis of sub-dataset I, sex is equally distributed, with females comprising 52\% (66 more females out of 2096 subjects) and males 48\% (Figure \ref{fig:glucagesex}). Likewise, the total number of measurements is similarly balanced, with females contributing 54\% and males 46\% (Figure \ref{fig:glucagesexfreq}). Females have up to 9.9 million more data points in total, which could potentially introduce a bias. Nevertheless, Figure \ref{fig:sexbox} reveals that the boxplots of both sexes are very similar, indicating that the glucose ranges do not differ much, with a similar median but a slightly increased mean in males. 
Figure \ref{fig:agedist} presents the age group distribution, with Subfigures \ref{fig:sexagef} and \ref{fig:sexagem} highlighting that adults aged 18 to 55 are the most representative. More females are included in these age groups, especially between 36 and 55 years. The second largest group comprises elderly adults ($\ge$56 years), with an equal sex distribution. The least present are newborns, children, and teenagers, also with a balanced sex ratio. The uneven distribution of subjects across age groups may pose a bias risk when analyzing the full cohort, whereas focusing on individual age groups may not include enough data to be fully representative.
Figures \ref{fig:glucagesexf} and \ref{fig:glucagesexm} also reveal that the ratio of glucose measurements varies between demographics, while most present are females between 36 and 55 years. Other age groups have a very similar distribution of values among sexes. Even if the group aged more than 56 has more subjects, the measurement frequency is similar for males but significantly less for females of the 36 and 55 years group. Children aged between 3 and 6 years, young adults (18-25 years), and adults (26-35 years) have a very similar distribution of glucose measurements, and also among sexes. The least representative are newborns (0-2) and children (11-13). 

Finally, examining bias possibility in more detail, Figure \ref{fig:agebox} presents the boxplots of the different age groups for the skewness of measured glucose values. While the variation is very similar for age groups between 26 and more than 56 years, differences can be noticed between newborns, children, teenagers, and adults. Hence, based on the glucose characteristics, the subjects can be grouped into 0-10 years, into 11-25 years, and into 26 to more than 56 years. The first group shows increased values with increased maximum, the second group has slightly decreased values, and the last group tends to have more controlled glucose ranges with less variance and smaller maximums. For the earlier ages until 10 years, more variance can be observed than for the adults and elderly, whereas the mean is nearer to the lower values. Conclusively, age groups can indeed propose a bias since glucose values show different behavior and patterns in each group. It is proposed to train a generalized model but also compare among children, teenagers and young adults, and adults and the elderly.

\subsection{Heart Rate Characteristics}

\begin{figure}[ht]
\centering
\begin{subfigure}{0.4\textwidth}
    \includegraphics[width=\textwidth]{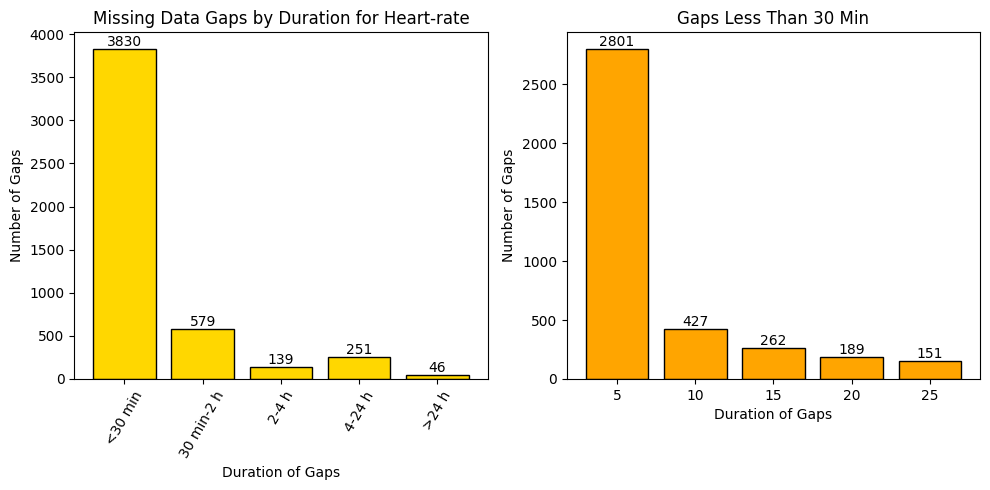}
    \caption{Missing Value Count of Sub-Dataset II}
    \label{fig:hrmissing}
\end{subfigure}
\begin{subfigure}{0.36\textwidth}
    \includegraphics[width=\textwidth]{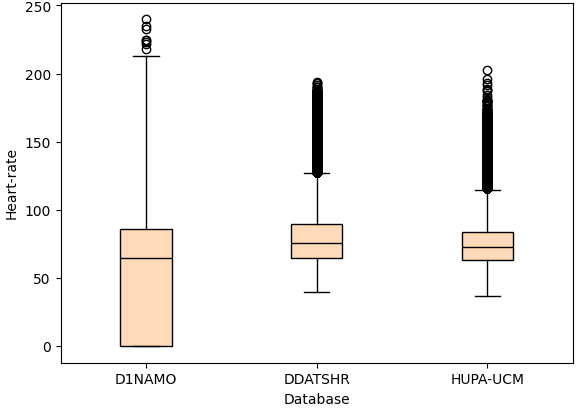}
    \caption{Variance of Heart Rate Data}
    \label{fig:hrbox}
\end{subfigure}
\caption{Analysis of Heart Rate Values}
\end{figure}

\begin{figure}[ht]
\centering
    \includegraphics[scale=0.3]{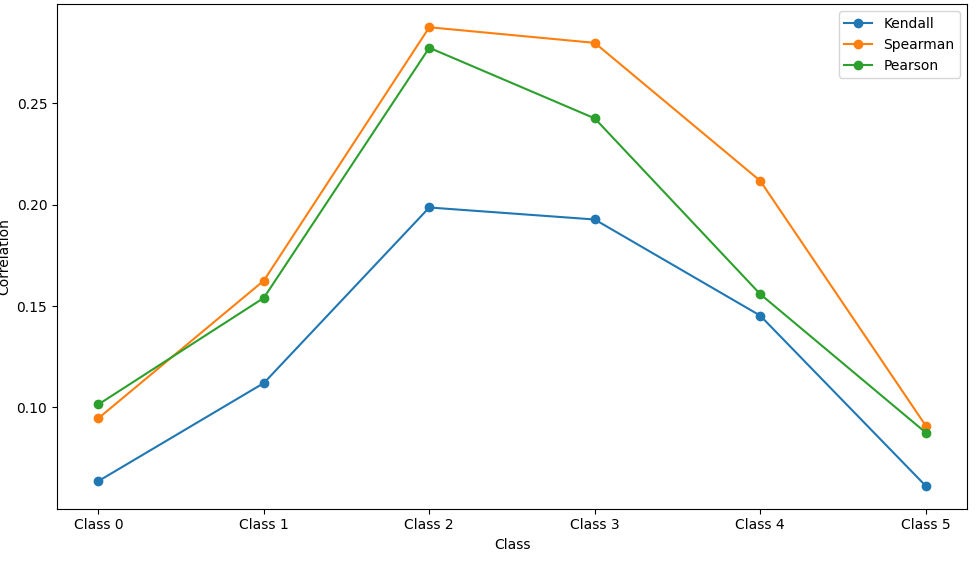}
    \caption{Correlation Between Hypoglycemic Glucose Values and Heart Rate Data}
    \label{fig:corrhr}
\end{figure}

Coming to sub-database II, missing values and the distribution of heart rate data are investigated. Similarly, most missing values are of short-term gaps of less than 30 minutes, while the majority of the gaps are of 5 minutes (Figure \ref{fig:hrmissing}). The boxplots show that fewer outliers exist while the variance is smaller than in the glucose measurements (Figure \ref{fig:hrbox}). The DDATSHR and HUPA-UCM datasets have similar variance, mean values, and minimums, while the variance and the maximum value of the DDATSHR dataset is increased in comparison. The majority of the values are between 80 and 200 beats per minute, while outliers are between 120 and 200 beats per minute. However, the D1NAMO database has a larger variance, which could be due to the difference in sensor brands, since a lot of sensor errors are captured as zeros rather than introducing them as missing values. These observations could irritate the boxplot, skewing it to lower bounds. In addition, the maximum value is above 210 beats per minute, and outliers are above this range, marking a great difference from the DDATSHR and HUPA-UCM datasets. This parameter could need further preprocessing, such as normalization and standardization, to enable the use of all datasets. 
\begin{table*}[h]
\centering
\caption{Correlation Between Glucose and Heart Rate Data}
\label{tab:corrhr}
\begin{tabular}{|c|c|c|c|c|c|c|c|c|}
\hline
\textbf{Correlation} &\textbf{Total}  &\textbf{Mean} & \textbf{Class 0} & \textbf{Class 1} & \textbf{Class 2} & \textbf{Class 3} & \textbf{Class 4} & \textbf{Class 5}   \\
\hline 
\textbf{Pearson} &0.0718 & 0.0362 & 0.1015
& 0.1540
&0.2773
&0.2425
&0.1559
&0.0877  \\ \hline

\textbf{Spearman} &0.0734& 0.0383
&  0.0947
&    0.1625
&    0.2874
&    0.2797
&   0.2117
&    0.0911 \\ \hline

\textbf{Kendall} &0.0494& 0.0262
&  0.0637
&  0.1120
&    0.1986
&   0.1926
&   0.1453
&  0.0614 \\ 
\hline
\end{tabular}
\end{table*}
Finally, the correlation between glucose and heart rate data was estimated with the Spearman's Rank correlation. The overall population correlation is 0.073, while the mean of all individual correlations per subject is only 0.038. The correlation of hypoglycemic points, assessed with different correlation methods, can be seen in Table \ref{tab:corrhr}. Class 0 stands for the hypoglycemic event defined as $\le70$ mg/dL, class 1 for 5-10 minutes before hypoglycemia, class 2 for 15-25 minutes before, class 3 for 30-55 before, class 4 for 60-120 before, and class 5 for no hypoglycemia risk within a short-term horizon. It can be observed that there is an increased moderate correlation between 30 min and 55 min, while hypoglycemia and no hypoglycemia are less related. For the correlation analysis, hypoglycemic data points were identified, and the time before the event was computed backward. More than 120 minutes before the event, were set to no hypoglycemia. Figure \ref{fig:corrhr} reveals the difference between the Pearson, Spearman's Rank, and Kendall correlations, visualizing the peak relation in class 2. It is noticed that there is a small decrease to class 3 in all correlations and then a significant decrease until class 5. These findings indicate that heart rate is a valuable feature to predict hypoglycemia, showing a positive relation with CGM values.

\section{Discussion}
\label{sec:discussion}

This study provides DiaData, an integrated dataset of 15 datasets collected from patients with T1D involving 149 million CGM values, demographics, and heart rate values. The dataset has an imbalanced distribution of glucose ranges, with in target values being the most prevalent, followed by hyperglycemic data points. Hypoglycemia comprises only 4\% and thus, is underrepresented, which could impact the performance of classification models or the forecasting capability of lower glucose ranges. Hence, depending on the intended application, unfiltered data may require the use of balancing methods. Moreover, separate models could be developed, tailored to specific glucose ranges, rather than being trained with all data. Heart rate data is also less present, with only 3 datasets included in sub-database II, highlighting the need for more multivariate datasets. While this sensor data is less collected, heart rate data was revealed as a valuable parameter that correlates with short-term hypoglycemia risk of 15-55 minutes before the adverse event. Notably, using both CGM values and heart rate data can improve and stabilize prediction performance. 

It was presented that the majority of the included datasets followed a similar data acquisition protocol and used CGM devices from the same manufacturers. Consequently, we assert that the integration of these datasets can be considered as feasible, enabling the training of population-based and individualized models with DiaData. Notably, DiaData encompasses a heterogeneous population, since the clinical trials often involved case and control groups and different age groups. Thus, subjects in the integrated cohort exhibited varying distributions of glycemic states, and were not limited to individuals with access to advanced CGM technology and software or increased awareness. Among the CGM devices utilized, Dexcom and FreeStyle Libre were the most prevalent, although different versions were employed, potentially contributing to variability in glucose estimation accuracy and precision. Devices of Metronic or the iPro2 Professional were used less frequently. Therefore, preprocessing and appropriate data scaling methods are essential to ensure the harmonization of measurements and to facilitate the effective use of DiaData.

Furthermore, a significant difference between the number of subjects and the measurement frequency of subjects was noticed across the single datasets before integration. In particular, the T1GDUJA dataset only comprises one subject. Thus, subjects with significantly more data points could rather be used for model individualization, in which generalized models can be retrained with transfer learning. Personalization can enhance model performance and deliver tailored decision support. This is particularly beneficial for small datasets limited to hypoglycemic values, since transfer learning can improve short-term predictions \cite{cinarMaster}.

Moreover, the quality of DiaData was analyzed. It was identified that both sexes are equally present in the integrated dataset and a variety of age groups are represented from newborns to the elderly, with an over-representation of adults ($\ge36$). Also, while glucose values are almost equally distributed among the sexes, variations can be identified across age groups. Different age groups show different variances of glucose levels. These could introduce a bias impacting the performance of ML models. The observed bias may also be due to the age groups that were defined, with younger cohorts divided into narrower intervals, and older cohorts grouped into wider age ranges. This imbalanced distribution can create inequalities within the demographic distribution. However, children grow very fast, and their biological requirements change very fast, which suggests a narrower differentiation. Based on the glucose characteristics and the boxplot analysis, a further classification into children, teenagers+young adults, and adults+elderly adults is suggested. 

Integrating all datasets has shown that different sensors and estimation methods make seamless integration of data challenging. Continuous sensor data causes a lot of missing values, which require proper imputation methods. Most missing values involve 5-minute gaps, which could have occurred due to sensor and environmental errors or undersampling. Moreover, since we have different gap sizes of consecutive missing values, those could require different imputation methods as stated in \cite{SuwenLin}, since glucose behavior is not linear. It is proposed to impute gaps up to 4 hours and remove the remaining gaps. Therefore, the dataset needs extensive preprocessing before being AI-ready. 

Improving the quality of the dataset and providing an AI-ready format serve as future aims. Then, models will be tested for hypoglycemia classification and glucose forecasting. Moreover, more feature can be extracted from the proposed datasets, and their correlation to hypoglycemic glucose values can be investigated.

\section{Conclusion}
\label{sec:conclusion}
This study integrated 15 databases collecting CGM data into DiaData, a large glucose database, which can be used to train ML or DL models to predict glucose levels or adverse events. A main database comprising 2510 patients was presented, which includes only CGM values. Two additional sub-databases can be derived from the main database, which either include demographics (10 databases with 2096 subjects) or heart rate measurements aligned with the CGM estimations (3 databases with 50 subjects). Characteristics of DiaData show that: 1) For the main database, the distribution among hyperglycemia, euglycemia, and hypoglycemia is imbalanced, with hypoglycemia and severe hypoglycemia being under-represented. Most missing values in each database are of short-term gaps of less than 30 minutes, in particular of 5 minutes, indicating sensor errors, misplacement, or sensor changes. 2) The distribution among sexes is equal in sub-database II, while the distribution across age groups differs. 3) Only 3 databases include heart rate values, however heart rate data correlates with glucose values 15-55 minutes before hypoglycemia. Future work will focus on removing outliers, enhancing data quality, and extracting more features.

\subsection*{Data and Code Availability}

The datasets used in this study were obtained from multiple third-party sources. Due to licensing restrictions, we do not redistribute the full integrated dataset. However, all preprocessing and integration scripts are publicly available at \href{https://github.com/Beyza-Cinar/DiaData}{https://github.com/Beyza-Cinar/DiaData}, along with instructions for obtaining the original datasets. A subset of the dataset can be downloaded from the following source: \cite{DiaDataZenodo}.

The sources of subsets of the data are the Barbara Davis Center, Jaeb Center for Health Research, Joslin Diabetes Center, T1D Exchange, University of Colorado, and the University of Virginia. Retrieved from: https://public.jaeb.org/dataset. The analyses, content, and conclusions presented herein are solely the responsibility of the authors and have not been reviewed or approved by the before mentioned institutions.

%
%
\printbibliography

\end{document}